\documentclass[11pt, a4paper, oneside]{article}
\usepackage[T1]{fontenc}
\usepackage[utf8]{inputenc}
\usepackage[dvipsnames]{xcolor}
\usepackage{bbm}
\usepackage{amssymb}
\usepackage{amsmath}
\usepackage{amsthm}
\usepackage{xargs} 
\usepackage{todonotes}
\usepackage{subfigure}
\usepackage{times}
\usepackage[a4paper]{geometry}
\usepackage{algorithm}
\usepackage{algpseudocode}
\usepackage{algorithmicx}
\usepackage{url}

\theoremstyle{definition}
\newtheorem{definition}{Definition}[section]
\theoremstyle{theorem}

\theoremstyle{remark}
\newtheorem{remark}{Remark}[section]
\theoremstyle{example}

\newcommand{\R}{\mathbb{R}}

\newcommand{\CC}{\mathcal{C}}

\newcommand{\ii}{\mathbf{i}}
\newcommand{\X}{\mathbf{X}}
\newcommand{\XX}{\tilde{X}}
\newcommand{\nn}{$n_{\text{left}}$}

\title{Band selection with Higher Order Multivariate Cumulants for small target
	detection in hyperspectral images}

\author{Przemysław Głomb\thanks{przemg@iitis.pl}, Krzysztof
	Domino\thanks{kdomino@iitis.pl}, Michał Romaszewski, Michał Cholewa\\
	Institute of Theoretical and Applied Informatics,\\
	Polish Academy of Sciences,\\
	Ba{\l}tycka 5, 44-100 Gliwice, Poland}
\date{\today}

\begin{document}
\maketitle

\begin{abstract}
In the small target detection problem a pattern to be located is on the order of magnitude less numerous than other patterns present in the dataset. This applies both to the case of supervised detection, where the known template is expected to match in just a few areas and unsupervised anomaly detection, as anomalies are rare by definition. This problem is frequently related to the imaging applications, i.e. detection within the scene acquired by a camera. To maximize available data about the scene, hyperspectral cameras are used; at each pixel, they record spectral data in hundreds of narrow bands -- providing in many cases enough information to e.g. identify individual materials or estimate their chemical degradation.

The typical feature of hyperspectral imaging is that characteristic properties of target materials are visible in the small number of bands, where light of certain wavelength interacts with characteristic molecules. Because of that, for small target detection the selection of bands where this interaction can be observed is the key. When given prior information about target spectrum, an algorithm could be optimized for its detection, however this problem is much more challenging when either that information is unavailable (anomaly detection) or requires additional human expert input (template spectrum is available without the context of chemical composition). A target-independent band selection method based on statistical principles is a versatile tool for solving this problem in different practical applications.

Combination of a regular background and a rare standing out anomaly will produce a distortion in the joint distribution of hyperspectral pixels. Higher Order Cumulants Tensors are a natural `window' into this distribution, allowing to measure properties and suggest candidate bands for removal. While there have been attempts at producing band selection algorithms based on the $3$\textsuperscript{rd} cumulant's tensor i.e. the joint skewness, the literature lacks a systematic analysis of how the order of the cumulant tensor used affects effectiveness of band selection in detection applications. In this paper we present an analysis of a general algorithm for band selection based on higher order cumulants. We discuss its usability related to the observed breaking points in performance, depending both on method order and the desired number of bands. Finally we perform experiments and evaluate these methods in a hyperspectral detection scenario.

\end{abstract}


\section{Introduction}

Hyperspectral imaging (HSI) systems simultaneously capture hundreds of narrow spectral channels, usually in the Visual-Near Infrared (VNIR, 400-1000nm) or Short Wave Infrared (SWIR, 1000-2500nm) regions of the electromagnetic spectrum. At each pixel the hyperspectral camera produces an approximation of a continuous spectrum. This data carries information about materials presents in the scene \cite{ghamisi2017advances}, as individual intensities result from interaction of light photons of given energy with different molecules.
HSI has many potential practical uses including
remote sensing of vegetation~\cite{thenkabail2016hyperspectral}, aiding in art conservation~\cite{grabowski2018automatic}, cultural heritage analysis~\cite{cucci2016reflectance}, forgery detection~\cite{rodionova2005nir} or gunpowder residue detection~\cite{glomb2018gunshot}.

Among many hyperspectral applications~\cite{bioucas2013hyperspectral}, a 
promising one is the target detection. Classic approaches to hyperspectral 
target detection~\cite{nasrabadi2014hyperspectral} include derivations of RX or 
SVDD detectors for unsupervised case and spectral matched filter (SMF) or 
subspace projections in the supervised case. In the latter case, the methods 
based on spectral angle (e.g. Spectral Angle Mapper) are also widely used and 
effective~\cite{pengra2007mapping}. These methods are often supported by 
algorithms from many domains, e.g., basic detectors are often used as a part of 
a more complicated algorithm that includes other Machine Learning approaches 
\cite{shi2014ship}, dedicated preprocessing and data-window schemes 
\cite{kwon2003adaptive}. Another group of methods is based on transforming the 
data representation, either by modelling the pixel-neighbours relation 
\cite{li2015collaborative} or by a sparse vector referring to the background 
spectra dictionary \cite{zhang2015sparse}. Hyperspectral band selection is also 
a common extension to existing methods, applied as preprocessing before classic detectors~\cite{sun2015new}.

While hyperspectral images are rich in features, their processing is
challenging because of the volume of each image (usually thousands times larger
than a corresponding RGB image) and the difficulty of obtaining ground
truth data, which requires expert knowledge and sometimes on-site
verification\cite{ghamisi2017advances}. Due to huge volume and high correlation
among the neighbouring bands, dimensionality reduction is often applied as a
preprocessing step to discard the redundant information in HSI data \cite{rodriguez2007unsupervised}.
A common approach is to apply feature extraction methods e.g. the Principal
Components Analysis (PCA) \cite{rodriguez2007unsupervised} in order to transform and reduce the dimensionality of the data. However, obtained features are not directly related to the original spectral channels, therefore an alternative approach is band selection~\cite{jia2016novel}, which selects a
subset of spectral channels that well represent the image according to some
criterion. The band selection has an advantage of preserving original
information about the spectrum~\cite{chang2006constrained}. This is important
e.g. in mineral analysis~\cite{brown2008marte} where spectral signatures of
minerals are expected to be associated with certain parts of the spectrum.

Band selection is an important component of many HSI processing methods, not limited to detection. Depending on whether the training data is available, band selection methods can be divided into supervised approaches that select bands based on training examples~\cite{li2012locality} and unsupervised methods~\cite{yuan2016dual}. While supervised methods can obtain more discriminative features, dependence on training examples may lead to instability of the solution, therefore unsupervised band selection may be more robust \cite{jia2016novel}. Unsupervised band selection approaches can be divided into clustering-based~\cite{martinez2007clustering}\cite{yuan2016dual} and ranking based~\cite{chang2006constrained}\cite{jia2016novel}. Clustering-based methods partition bands into disjoint groups (clusters) such that bands in the same cluster are similar to each other and dissimilar to the rest, then the most representative band from each cluster is selected~\cite{yuan2016dual}. One particular motivation for this approach may be to overcome the complexity in calculating the joint distribution in high-dimensional spaces~\cite{martinez2007clustering}. Ranking-based methods assess the importance of each band according to a certain criterion, such as non-Gaussianity~\cite{chang2006constrained} or volume-gradient~\cite{geng2014fast}. Recently, methods that try to combine ranking-based and clustering approaches were also proposed~\cite{jia2016novel}.

Modelling the data distribution with multivariate Gaussian distribution has
many successful application in pattern recognition, e.g.,
face~\cite{turk1991eigenfaces} and gesture~\cite{gawron2011eigengestures} classification
or feature selection~\cite{sheffield1985selecting}. Hence a number of existing
classification and target detection algorithms are based on the multivariate
Gaussian model, however in many cases this model does not represent the
statistical behaviour of hyperspectral data~\cite{acito2007statistical}. This
motivates the use of non-Gaussian model for HSI data processing
\cite{kajic2010robust}. It is also well-known that data can have a non-Gaussian
joint distribution despite Gaussian marginals \cite{domino2018use}. However,
while
joint distribution is important for HSI analysis, it is difficult to estimate
\cite{martinez2007clustering}, therefore approaches based on copulas are
employed instead \cite{zeng2009band}.

In our work we use cumulants of multivariate data for band selection. Those 
cumulants, as discussed in the introduction of~\cite{domino2018efficient}, are 
a `window' to the joint distribution of multivariate data. The second cumulant 
of multivariate data is simply its covariance matrix. It can be used in a 
feature selection procedure e.g. by recurrently removing least informative 
features in such way that the volume of the information hyper-ellipsoid of the 
covariance matrix of remaining features is maximised. This leads to the Maximum 
Ellipsoid Volume (MEV) feature selection \cite{sheffield1985selecting}. 
However, since the MEV is based on the multivariate Gaussian distribution of 
data, parametrised by the covariance matrix, it is not sensitive on 
non-Gaussian features such as skewness or kurtosis~\cite{geng2014fast}.
Higher Order Multivariate Cumulants (HOMC) are represented in our work by
tensors of order
$d$
($d$-dimensional arrays)~\cite{kolda2009tensor}. These tensors have an
important property: when data has a multivariate Gaussian distribution, every
element of a cumulant tensor of order $d\ge3$ equals
zero~\cite{kendall1946advanced,lukacs1970characteristic}. This property can be
utilised when searching for non-Gaussian distributed bands, and leads to
creation of the Joint Skewness Band Selection (JSBS)~\cite{geng2015joint}
procedure, where a Higher Order Singular Value Decomposition
(HOSVD)~\cite{de2000multilinear} of a $3$-order cumulant's tensor is used to
create a measure of non-Gaussianity and select the most informative bands.

However, we have observed that the higher the cumulant's tensor order $d$ is, the
more it is sensitive to tails in multivariate distribution, where outliers may
appear. This is due to the fact, that while estimating $d$\textsuperscript{th}
order cumulant we
have terms proportional to the difference of the sample from the mean raised to
the $d$\textsuperscript{th} power. This problem was discussed in \cite{muzy2001multifractal} where multivariate
financial data analysis is performed and this type of outlier is associated with an anomalous situation (e.g. a financial crisis).
The observation regarding the effect of HOMC motivates a proposal of a family of
methods in which a $d$-order cumulant's tensor is used to extract information
about the non-Gaussian joint distribution of features from hyperspectral data.
In addition, we also propose a general method of normalisation that reproduces
the module of asymmetry or kurtosis in an univariate case. Further as
discussed in \cite{geng2015joint} for the particular $3$\textsuperscript{rd}
cumulant case such normalisation reduces cross-correlation between chosen
features. This effect have appeared as well in a general $d$\textsuperscript{th}
cumulant case during our analysis.

In this paper we perform an experimental evaluation of band selection methods based on
HOMC for small target detection in hyperspectral images. Specifically, we make
the following contributions:
\begin{enumerate}

\item We apply Joint Kurtosis Feature Selection (JKFS) \cite{domino2018use} to the problem of band selection and show that it can be effectively applied for hyperspectral small target detection.

\item We introduce a new method of band selection, called Joint Hyper Skewness
Feature Selection (JHSFS) that is the extension of JKFS. We discuss its
properties and show that in some hyperspectral detection scenarios, the
proposed method can outperform both JSBS and JKFS.

\item We propose a uniform derivation of $d$-order cumulant-based band selection methods, that
derives JSBS (order $d=3$), JKFS (order $d=4$) and JHSFS (order $d=5$) as
special cases and can be extended to orders $d>5$. We also discuss both
advantages and disadvantages of HOMC-based methods.

\item We present a comparison of performance evaluation for cumulant-based
methods on real-life hyperspectral data. In addition, we make an
important observation that the performance of band selection methods based on
HOMC depends on both the method's order and the
desired
dimensionality of the feature space (i.e., the number of selected
bands) and can drop sharply for small number of bands. Hence every HOMC-based
method of band selection has a minimum required number of selected bands. The
existence of this
phenomena, called `breaking points', is shown in our experimental evaluation.
We propose an explanation by providing a hypothesis
about its relation to the number of off-diagonal elements in $d$-order tensor.
\end{enumerate}

\section{Band selection using Higher Order Multivariate Cumulants}

In this section we focus on the formal introduction of feature selection based
on Higher Order Multivariate Cumulants (HOMC). As stated in the introduction, we focus on the information carried by
non-Gaussian features.

\subsection{Preliminaries}

We use the bold uppercase notation for a matrix (e.g. $\X$) and an uppercase for its column vector (e.g. the $i$\textsuperscript{th} vector of matrix $\X$ is $X_i$). We use the lowercase
notation for an element of a matrix or a tensor (e.g. $x_{ij}$).  We introduce tensors with the following definition.

\begin{definition}
Following \cite{kolda2009tensor}, a
$d$-mode
tensor
\begin{equation}
\mathcal{A} \in \R^{n_1 \times \ldots
\times n_d},
\end{equation} is a multidimensional array with elements $a_{\ii}$ indexed by the
multi-index $\ii = (i_1, \ldots, i_d)$, where $i_1 \in [1,2,\ldots, n_1],
\ldots, i_d \in [1,2,\ldots, n_d]$.
\end{definition}
 Following the Chapter $2.4$ in
\cite{kolda2009tensor}, a tensor unfolding or matricisation is a
transformation of a tensor
into a matrix by integrating $d-1$ modes of the tensor into one mode of the
new matrix, and is formally defined as follows:

\begin{definition}\label{def::unf}
	Following \cite{kolda2009tensor}, the $k$-mode unfolding of a tensor
	$\mathcal{A}
	\in \R^{n_1 \times \ldots \times
	n_d}$
	into a matrix
	\begin{equation}
		\mathcal{A}_{(k)} = \mathbf{B} \in R^{n_k \times
			t} \text{ where } t = \prod_{r = 1, r \neq k}^d n_r,
	\end{equation} is defined element-wisely as:
\begin{equation}
	b_{i_k, j} = a_{i_1, \ldots, i_k, \ldots, i_d},
\end{equation}
where
\begin{equation}j = 1+\sum_{l = 1 \ l\neq k}^d\left(i_l-1\right) \prod_{r = 1
\ r \neq k}^{l-1} n_r.
\end{equation}
\end{definition}

\subsection{HOMC for hyperspectral data}\label{sub::hochd}
The hyperspectral image data can be represented in the form of a $3$-mode tensor:
\begin{equation}
\mathcal{X} \in \R^{p_x \times p_y	\times n},
\end{equation}
where first two modes correspond to spatial dimensions of pixels while last mode
correspond to spectral channels.
In this paper we are searching for spectral channels
that carry relevant information about small targets i.e. they are present in small fraction of pixels. Following \cite{geng2015joint} we consider spectral channels as
marginals of a multivariate variable and reflectance values recorded at each given pixel as a realisation of such variable.
Since we are analysing multivariate statistics of data, we are ignoring spatial information in data (pixel positions). Therefore for clarity we unfold hyper-spectral data, in accordance with the Definition~\ref{def::unf}, into the
matrix form, where rows indicate realisations (hyperspectral pixels) and columns indicate marginals
(spectral channels), as:
\begin{equation}\label{eq::datametrix}
\mathbf{X} =  \left(\mathcal{X}_{(3)}\right)^{\intercal} \in \R^{t \times n},
\end{equation}
where $t=p_x p_y$.
Consistently with notation introduced in
\cite{domino2018efficient}, a column vector of all realisations for
$i$\textsuperscript{th} spectral channel is $X_i \in \R^t$. A conceptually
simple approach to search for a
sub-set of non-Gaussian marginals would be
to test each univariate $X_i$, indexed by $i = 1,\ldots, n$, for normality.
However as
discussed in \cite{geng2015joint}, such approach is oversimplified since
it does not take into account multi-variate non-Gaussianity.
This can be explained using a reference to a copula
approach~\cite{nelsen1999introduction}, because
data can have Gaussian marginal distributions and non-Gaussian copula
(non-Gaussian cross correlation between marginals). This problem is
described in detail in ~\cite{domino2018use}.

A HOMC can be represented by super-symmetric
tensors~\cite{schatz2014exploiting}, defined as follows:

\begin{definition}
The $d$-mode tensor $\mathcal{A}
\in \R^{n_1 \times \ldots \times
	n_d}$ of size $n_1 = \ldots = n_d = n$ is super-symmetric, if values of its
	elements are invariant under any permutation $\pi$ inside its multi-index
	\textit{i.e.}
\begin{equation}
\forall_{a_{\ii} \in \mathcal{A}} \forall_{\pi} \  a_{\ii} = a_{\pi(\ii)}.
\end{equation}
Referring to~\cite{domino2018efficient} we denote the
super-symmetric tensor as $\mathcal{A} \in \R^{[n,d]}$.
\end{definition}

\begin{remark}\label{re::symunfold}
	For the super-symmetric tensor $\mathcal{A} \in
	\R^{[n,d]}$ all unfoldings (see the Definition~\ref{def::unf}) are equivalent,~\textit{i.e.}
	\begin{equation}
	\mathcal{A}_{(1)} = \ldots = \mathcal{A}_{(d)},
	\end{equation}
	since a multi-index of a super-symmetric tensor can be permutated without changing a value of indexed element.
\end{remark}

\begin{definition}\label{eq::expvalue}
Given data matrix $\mathbf{X} \in \R^{t
\times n}$ with elements $x_{j,i}$
where $j$ indexes realisations (hyperspectral pixels) and $i$ marginals (spectral channels), we define the following
estimators of expected values~\cite{domino2018efficient}:
\begin{equation}
E(X_{i_1}, \ldots X_{i_d}) = \frac{1}{t} \sum_{j=1}^t \prod_{k=1}^d x_{j, i_k}.
\end{equation}
Analogically centred expected values~\cite{domino2018efficient} are defined as:
\begin{equation}
E(\XX_{i_1}, \ldots, \XX_{i_d}) = E\left(\left(X_{i_1} -
E(X_{i_1})\right),
\ldots, \left(X_{i_d}-E(X_{i_d})\right)\right).
\end{equation}
\end{definition}
We assume that the number of realisations $t$
is relatively large, therefore the bias of the estimator is negligible.
In addition, we note that expected values are invariant to any permutation of
realisations as long as it is performed simultaneously for all marginals.

Referring to~\cite{domino2018efficient}, a $d$\textsuperscript{th} order
cumulant of $n$-variate data is a super-symmetric tensor $\CC_d \in
\R^{[n, d]}$ and in particular:
\begin{enumerate}
	\item $\CC_1 \in \R^n$ with elements $c_i = E(X_i)$,
	\item $\CC_2 \in \R^{[n,2]}$ with elements $c_{i_1, i_2} =
	E(\XX_{i_1},\XX_{i_2})$,
	\item $\CC_3 \in \R^{[n,3]}$ with elements $c_{i_1, i_2, i_3} =
		E(\XX_{i_1},\XX_{i_2},\XX_{i_3})$,
	\item $\CC_4 \in \R^{[n,4]}$ with elements
	\begin{equation}\label{eq::c4}
	\begin{split} c_{i_1, i_2, i_3, i_4} &=
			E(\XX_{i_1},\XX_{i_2},\XX_{i_3},\XX_{i_4}) -
			E(\XX_{i_1},\XX_{i_2})E(\XX_{i_3},\XX_{i_4}) \\ &-
			E(\XX_{i_1},\XX_{i_3})E(\XX_{i_2},\XX_{i_4}) -
			E(\XX_{i_1},\XX_{i_4})E(\XX_{i_2},\XX_{i_3}).
			\end{split}
	\end{equation}
	\item $\CC_5 \in \R^{[n,5]}$ with elements
	\begin{equation}\label{eq::c5}
	\begin{split} &c_{i_1, i_2, i_3, i_4, i_5} =
	E(\XX_{i_1},\XX_{i_2},\XX_{i_3},\XX_{i_4},\XX_{i_5}) \\ &-
	\underbrace{E(\XX_{i_1},\XX_{i_2})E(\XX_{i_3},\XX_{i_4},\XX_{i_5}) -
	E(\XX_{i_1},\XX_{i_3})E(\XX_{i_2},\XX_{i_4},\XX_{i_5}) - \ldots}_{\times 10}
	\end{split}.
	\end{equation}
\end{enumerate}
Formulas for cumulants of order $d>5$ are out of scope of this paper, for their definition please refer to to~\cite{domino2018efficient}.

\begin{remark}\label{re::offdiag}
	Let $\CC_d$ be the cumulant's tensor of order $d \geq 3$ with elements
	$c_{\ii}$. Such tensor can be analysed by dividing it into the following disjoint
	areas:
	\begin{enumerate}
		\item diagonal elements $a_{\ii}$, where all elements of the
		multi index are the same, i.e.,~$i_1 = i_2 = \ldots = i_d = i$, such 
		area
		corresponds to univariate
		cumulants of marginals (wavelength in our case) numerate by $i$;
		\item off diagonal elements $a_{\ii}$, where all multi index elements
		are distinct~i.e.,~$i_1 \neq i_2 \neq \ldots \neq i_d$;
		\item partially diagonal elements $a_{\ii}$ that are neither
		diagonal nor off-diagonal~i.e.,~$\exists_{j,k,l}:
		i_j = i_k \neq i_l$.
	\end{enumerate}
\end{remark}

We make an important observation that off-diagonal elements of $d$-order
cumulant's tensor are tied to a mutual cross-correlation of $d$ marginals
(spectral
channels), therefore they give a new type of information for hyperspectral data
analysis. As an example, consider the $4$\textsuperscript{th} cumulant.
If its off diagonal elements fulfil $c_{i_1,i_2, i_3, i_4} = 0$ then
according to the Equation~\eqref{eq::c4}, the $4$\textsuperscript{th} central moment
$E(\XX_{i_1},\XX_{i_2},\XX_{i_3},\XX_{i_4})$ can be expressed in terms of
covariances ($2$\textsuperscript{nd} central moments) and carries no
additional information. Hence considering mutual cross-correlation of $4$
marginals would give no additional information. This is not the case if
$c_{i_1,i_2, i_3, i_4} \neq 0$, since we obtain additional information from
mutual cross-correlation of $4$ marginals.

In a case of the $3$\textsuperscript{rd} cumulant, off diagonal elements of
$c_{i_1,i_2, i_3}$ measure a specific type of mutual asymmetry of $3$ distinct
marginals. Analogically in the case of the $5$\textsuperscript{th} cumulant 
(see Equation~\eqref{eq::c5}) off
diagonal elements are much more complicated. Nevertheless one
can argue that they measure information that is not measured by 
$2$\textsuperscript{nd} and $3$\textsuperscript{rd} cumulants.

\subsection{Applications of HOMC to band selection}

Recall the standard
Maximum Ellipsoid Volume (MEV) method of feature selection
\cite{sheffield1985selecting} where at each iteration step, one marginal
variable is removed in such a way that the determinant of the covariance matrix
(second cumulant) is maximised. Such determinant is a product of
eigenvalues, hence proportional to the volume of the information ellipsoid. Inspired by this approach, authors of \cite{geng2015joint} introduced a method called JSBS (Joint Skewness Band Selection). They argued that by analogy, the determinant of the following matrix:
\begin{equation}
\label{eq:jsbs_mat}
\R^{[n, 2]} \ni \mathbf{M}_3 = \left(\CC_3 \right)_{(1)} \left( \left(\CC_3
\right)_{(1)} \right)^{\intercal},
\end{equation}
measures the information extracted by the $3$\textsuperscript{rd} cumulant
tensor -- $\CC_3$. Note that the Eq.~\eqref{eq:jsbs_mat} uses the first mode
unfolding, because by the Remark~\ref{re::symunfold}, the super-symmetric
tensor unfolding is mode invariant.
Based on this assumption, they introduced the
target function:
\begin{equation}
\label{eq:jsbs_fun}
f_{JSBS} = \frac{\sqrt{\det(\mathbf{M}_3)}}{\left(\det
\CC_2\right)^{\frac{3}{2}}},
\end{equation}
Band selection is then performed, analogically to MEV algorithm, by iteratively
removing marginal variables in such way that in every iteration a target function is
maximised. The denominator of the Eq.~\eqref{eq:jsbs_fun} is a normalisation
that, according to~\cite{geng2015joint}, reduces the risk of selecting
highly correlated marginals.

However, as mentioned in the introduction, we observed that the higher the cumulant's tensor order $d$ is, the more it is sensitive to tails in multivariate distribution, where outliers may appear. In addition, there exist datasets for which
$4$\textsuperscript{th} cumulant tensor generalisation, called Joint
Kurtosis Feature Selection (JKFS)
~\cite{domino2018use} is more effective that
JSBS.
This motivates an introduction of a general, $d$-cumulant based method.

We define a $d$-order dependency matrix as
\begin{equation}\label{eq::hosvd}
\R^{[n, 2]} \ni \mathbf{M}_d = \left(\CC_d \right)_{(1)} \left( \left(\CC_d
\right)_{(1)} \right)^{\intercal}.
\end{equation}
We use this matrix to define the following target function:
\begin{equation}\label{eq::targetf}
f_d = \frac{\sqrt{\det(\mathbf{M}_d)}}{\left(\det
	\CC_2\right)^{\frac{d}{2}}}.
\end{equation}

In the  case of $d = 3$ our methods simplifies to
JSBS, while in the case of $d = 4$ it simplifies to JKFS. As a
$5$\textsuperscript{th} cumulant is called hyper-skewness, in the case of $d=5$ we call our method the Joint Hyper Skewness Feature Selection (JHSFS).

The power term $\frac{d}{2}$ in the denominator provides the scale-invariant
normalisation. Suppose we have data $\X\in\R^{t \times n}$ and rescale all realisations of all
marginals by a factor $\alpha$~i.e.,~$\R^{t \times n} \ni \mathbf{Y} = \alpha \X
\in \R^{t \times n}.$ Referring to the general formula for the
$d$\textsuperscript{th} order multivariate cumulant
in~\cite{domino2018efficient}, and Equation~\eqref{eq::hosvd}, we have:
\begin{equation}
\begin{split}
\CC_d(\mathbf{Y}) &= \alpha^d \CC_d(\X), \\
\mathbf{M}_d(\mathbf{Y}) &= \alpha^{2d} \mathbf{M}_d(\X), \\
\det(\mathbf{M}_d(\mathbf{Y})) &= \alpha^{2dn} \mathbf{M}_d(\X), \\
f_d(\mathbf{Y}) &= \frac{\alpha^{dn}}{\alpha^{2dn/2}}f_d(\X) = f_d(\X).
\end{split}
\end{equation}
Another argument for such normalisation is that if $\X$ is a single
column of realisations of one marginal, it is easy to show that $f_3$ would be
an absolute value of its asymmetry while $f_4$ would be an absolute value of
its kurtosis. Finally, as argued in~\cite{geng2015joint}, such normalisation
can decrease the probability of choosing two highly cross correlated
features.

While the use of HOMC is sensitive to outliers located in tails of
multi-variate
distribution, it comes at a cost of higher estimation error. The theoretical
discussion regarding this estimation errors is provided in the Appendix~$A$
in~\cite{domino2018efficient}.

\subsection{Generalised band selection with HOMC}
To define a generalised band selection with HOMC we first introduce a fiber cut
of the super-symmetric tensor in the Definition~\ref{def::fcut}. The proposed
band selection method is then presented in the
form of the Algorithm~\ref{alg::itterremove}.

\begin{definition}\label{def::fcut}
	Let $\CC_d \in \R^{[n, d]}$ be the super-symmetric tensor. Following
	\cite{geng2015joint}, we define its $r$\textsuperscript{th} fibres cut,
	by the following tensor
	${\CC_d}_{(-r)} = \CC_d' \in \R^{[(n-1), d]}$, where:
	\begin{equation}
	c'_{i'_1, \ldots i'_d} = c_{i_1, \ldots, i_d}: i'_k = \begin{cases} i_k
	\ &\text{if} \  i_k < r \\
	i_{k}-1
	\ &\text{if} \ r \leq i_k < n,
	\end{cases}
	\end{equation}
	and $i_k' \in (1:n-1)$. Using notation from \cite{kolda2009tensor} we
	remove all $r$\textsuperscript{th} fibres of $\CC_d$. We note that such
	transformation preserves super-symmetry.
\end{definition}

 \begin{algorithm}[h]
 	\caption{Generalised, HOMC-based band selection algorithm
 		\label{alg::itterremove}}
 	\begin{algorithmic}[1]
 		\State \textbf{Input}: covariance $\CC_{2} \in 
 		\R^{[n, 2]}$, cumulant tensor $ \CC_{d} \in
 		\R^{[n, d]}$, target function $f_d$,
 		retained number of bands \nn $\le n$
 		\State \textbf{Output:} a subset (index) of bands
 		that carry meaningful information.
 		\Function{features select}{$\CC_{2}, \CC_{d}, f_d, n_{\text{left}}$}
 		\For {$n' \gets n$ to \nn}
 		\For {$i \gets 1 \textrm{ to } n'$}
 		\State $m_i = f_d\left({\CC_{2}}_{(-i)}, {\CC_{d}}_{(-i)}\right)$
 		\EndFor
 		\State set $r: m_r = \text{max}(\{m_1, \ldots, m_{n'}\})$ \hspace*{1cm} \(\triangleright\) remove band $r$
 		\State $\CC_{2} = {\CC_{2}}_{(-r)}$
 		\State $\CC_{d} = {\CC_{d}}_{(-r)}$
 		\EndFor
 		\State \Return remaining $n_{\text{left}}$ bands
 		\EndFunction
 	\end{algorithmic}
\end{algorithm}

\subsection{Theoretical limits for band selection with
HOMC.}\label{sec::parran}
Band selection procedure presented in the Algorithm~\ref{alg::itterremove} is
parametrised by a stop condition $n_{\text{left}}$ indicating the number of retained
features. It can be argued that this stop condition parameter coincides
with the used cumulants' order parameter $d$.
For the typical experimental setting we start with $n
= 50$ features.

The features elimination method we introduced first discards Gaussian
distributed spectral channels that contain either a pure noise or a background
Gaussian distributed marginals. For the subset of such marginals all HOMC are
zero. At this stage our elimination method removes the same bands regardless of cumulant's order.

\begin{figure}
	\centering
	\includegraphics{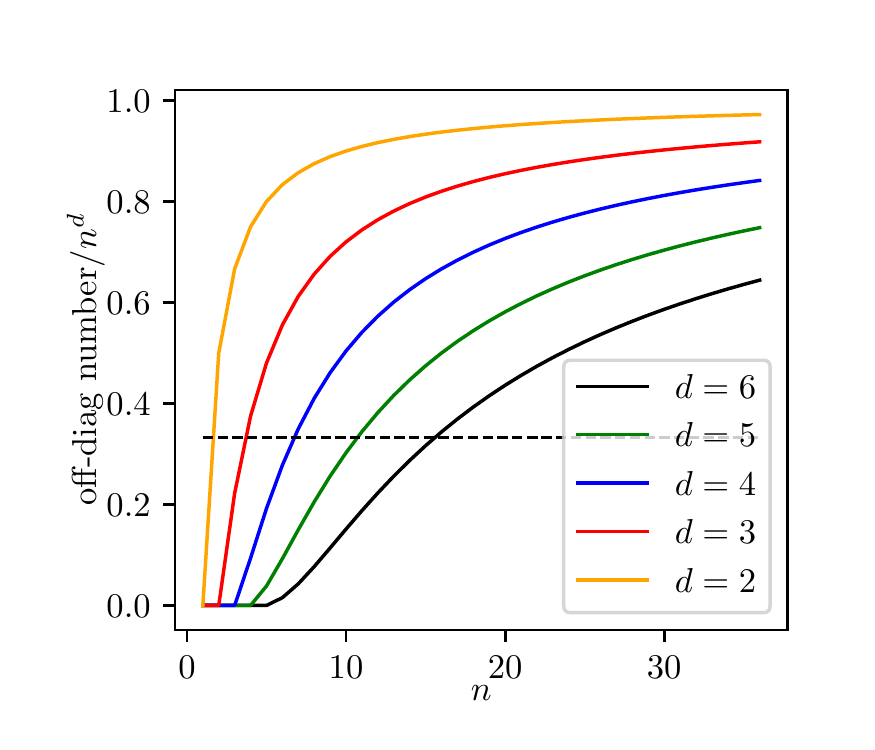}
	\caption{The ratio of off diagonal to all elements in a $d$-mode cumulant's 
	tensor as a function of feature vector size $n$. The dashed line marks the 
	$\frac{1}{3}$ ratio, below which the impact of off diagonal elements is 
	small (see Section~\ref{sec::parran}).}\label{fig::offdiag}
\end{figure}

Recalling Remark~\ref{re::offdiag} there are three areas of the cumulant's
tensor: diagonal, partially diagonal and off diagonal one. Following discussion
in Section~\ref{sub::hochd} we are particularly interested in an off diagonal
one since there an information about mutual cross-correlation of $d$
marginals may appear. This makes cumulants of higher order especially interesting. The fraction of off diagonal elements in $\CC_d \in \R^{[d,n]}$ is presented in
Figure~\ref{fig::offdiag}. This fraction drops sharply both with $n$ and $d$.
As a rule of thumb, we approximate that the impact of the off diagonal area in an algorithm is small if
its size (off diagonal elements numbers) is smaller that $\frac{1}{3}$ of the
tensor size. Referring to the Figure~\ref{fig::offdiag} it appears if
$n_{\text{left}} < 4$ for $d = 3$, $n_{\text{left}} < 7$ for $d = 4$,
$n_{\text{left}} < 11$ for $d = 5$ and $n_{\text{left}} < 16$ for $d = 6$.
Hence we expect those to be lower limits of features selection usability.

Consider at this point, that for large $d$ we may have large estimation error
of HOMC especially dealing with the moderate sample size i.e.,~$t
\approx 3\cdot 10^5$. Recall that an estimation error of high
order statistics grows rapidly with their order $d$, see Appendix $A$ in
\cite{domino2018efficient}. For this reason despite good detectability of the
JSBS~\cite{geng2015joint} and the JKFS~\cite{domino2018use} for such datasets
we may expect reduced performance for orders $d > 4$.

\subsection{Hyperspectral small target detection with HOMC}

The detection proceeds in two stages. First the band selection is performed
using the Algorithm~\ref{alg::itterremove} or the MEV, hence reducing the $p_x
\times
p_y \times n$ image to $p_x \times p_y  \times $ \nn\ image, then a detector is
applied to the selected bands to mark discovered targets.

For the detector we use the Spectral Angle Mapper (SAM)~\cite{Kruse1993SAM}. The SAM similarity score is computed between the elements $x_{j,i}$ of data matrix $\mathbf{X} \in \R^{t \times n}$ and the target signature matrix  $\mathbf{S} \in \R^{1 \times n}$ with the elements $s_{i}$ by applying the equation
\begin{equation}
d_{SAM} (\mathbf{X},j, \mathbf{S})=
\mathrm{cos}^{-1} \left(
	\frac{\sum_{i=1}^{n_{left}} x_{ji} s_i
	}{\sqrt{\sum_{i=1}^{n_{left}} x_{ji}^2}\sqrt{\sum_{i=1}^{n_{left}} s_i^2}}
\right).
\end{equation}
This detector has a number of advantages: it is simple to compute and has a straightforward explanation (it is sensitive to the angle between $n$-dimensional spectral vector, robust to e.g. illumination changes); it does not require to estimate additional parameters (besides a detection threshold); it does not require assumptions on data distributions (e.g. a normality assumption, present in some methods, which is rarely satisfied for real-life hyperspectral dataset); it has been well examined in many case studies, e.g. \cite{vanderMeer1997lithologies,Yao2010aflatoxin,Hunter2002Thames,Petropoulos2010burnt}.

The detection threshold, that is the only parameter of the detector, is required for evaluation of the boolean expression $ d_{SAM}
(\cdot) \geq \tau$ which decides whether a given spectral vector
$j$ is classified as a target or not. Given a detector described by the
Receiver
Operating Characteristic (ROC) curve, the parameter $\tau$ controls the balance between expected true and false positive
rates. This value can be optimized e.g. by cross-validation or by analysis of
distribution. We note that for comparison of the effect of different orders $d$
this parameter is not required, as whole ROC curves can be used for this task.

\section{Experimental results and discussion}

This section presents an experimental evaluation of band selection methods
based on Higher Order Multivariate Cumulants (HOMC) discussed in previous sections. The experiments are performed for
three methods of band selection based on HOMC: JSBS (order $d=3$), JKFS (order
$d=4$), and JHSFS (order $d=5$). As a reference, detection results for band
selection with Maximum Ellipsoid Volume (MEV), and without band selection are
also provided. Results of experiments are presented in the form of Receiver
Operating Characteristic (ROC) curves and the performance of the detector is
measured using the Area Under Curve (AUC) measure. Each ROC curve represents
the whole range of the detector parameter $\tau$, for a set number of retained
bands $n_{\text{left}}$, selected with the Algorithm~\ref{alg::itterremove}.

\subsection{Hyperspectral dataset}

\begin{figure}
	\subfigure[The Cuprite image (visualized with bands $1989\mathrm{nm}$, $2139\mathrm{nm}$ and $2288\mathrm{nm}$ as R, G, B to enhance the surface mineral composition)]{\includegraphics[width =
	0.48\textwidth]{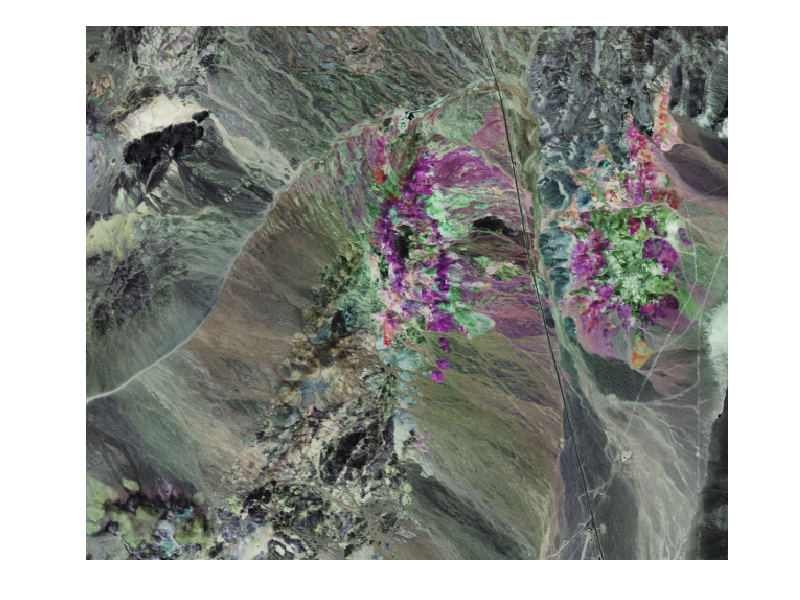}}
	\subfigure[Buddigntonite ground truth mask based on the maps from 	\cite{Swayze2014Cuprite}]{\includegraphics[width =
	0.48\textwidth]{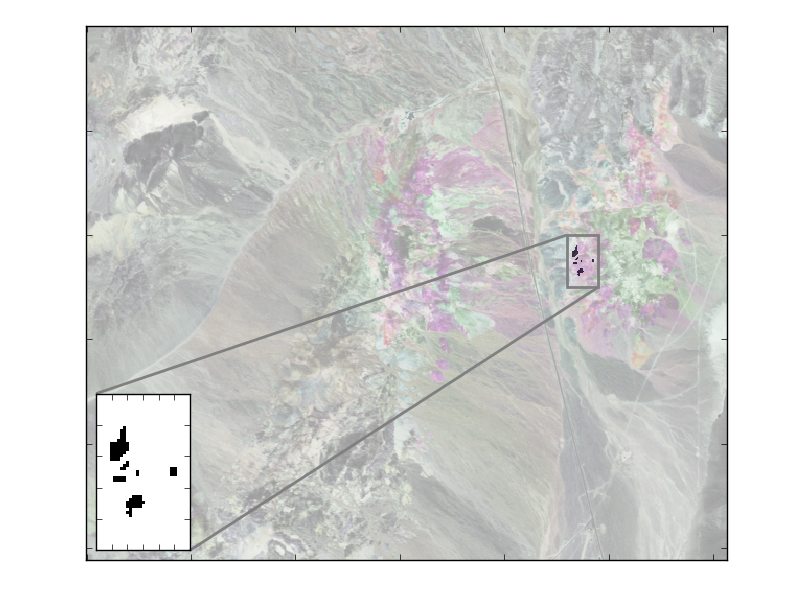}}
	\caption{Cuprite hyperspectral dataset and the ground truth used in detection experiments.}\label{fig::cuprite}
\end{figure}

In order to present the detailed examination of the performance of the proposed method, we use the Cuprite\footnote{Available online at \url{http://aviris.jpl.nasa.gov/data/free_data.html}.} hyperspectral dataset. The image presents a mining area around Cuprite in Las Vegas, NV, U.S. The site was imaged with AVIRIS sensor and has 224 spectral channels in range 370 nm to 2480 nm. In conformance with a standard procedure, noisy and water absorption bands 1–3, 104–113, 148–167, and 221–224 were removed from the image. To reduce computational complexity, we consider last 50 bands that contain the spectral range of interest~\cite{Dobigeon2009jointBayesian,Clark2003Earth}. The site has been a subject of many experiments, and its geology was mapped in detail \cite{Swayze2014Cuprite}.

In order to compare our results with these provided in~\cite{geng2015joint}, we focus on Buddigntonite deposit detection, which in Cuprite image has known local surface presence around the area nicknamed the `Buddigntonite bump'. Based on the \cite{Swayze2014Cuprite}, a ground truth map was prepared (see Figure~\ref{fig::cuprite}). To achieve independence of the target spectrum from the image, we use the corresponding reference spectrum\footnote{s07\_AV97\_Buddingtnt+Na-Mont\_CU93-260B\_NIC4b\_RREF} from the USGS Spectral Library~\cite{Clark2007SpectralLib}.

\begin{figure}
	\subfigure[]{\includegraphics[width =
		0.48\textwidth]{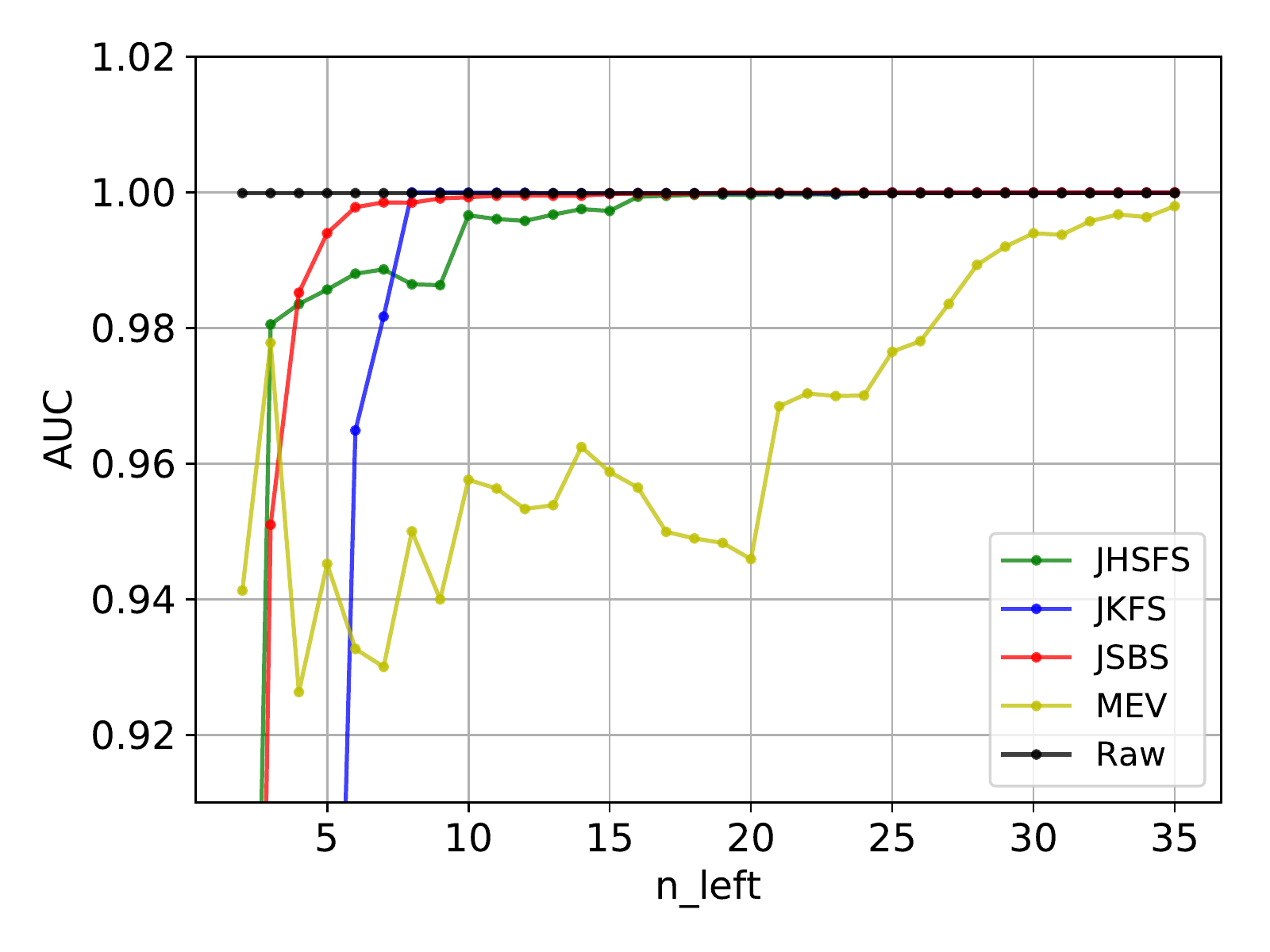}}
	\subfigure[]{\includegraphics[width =
		0.48\textwidth]{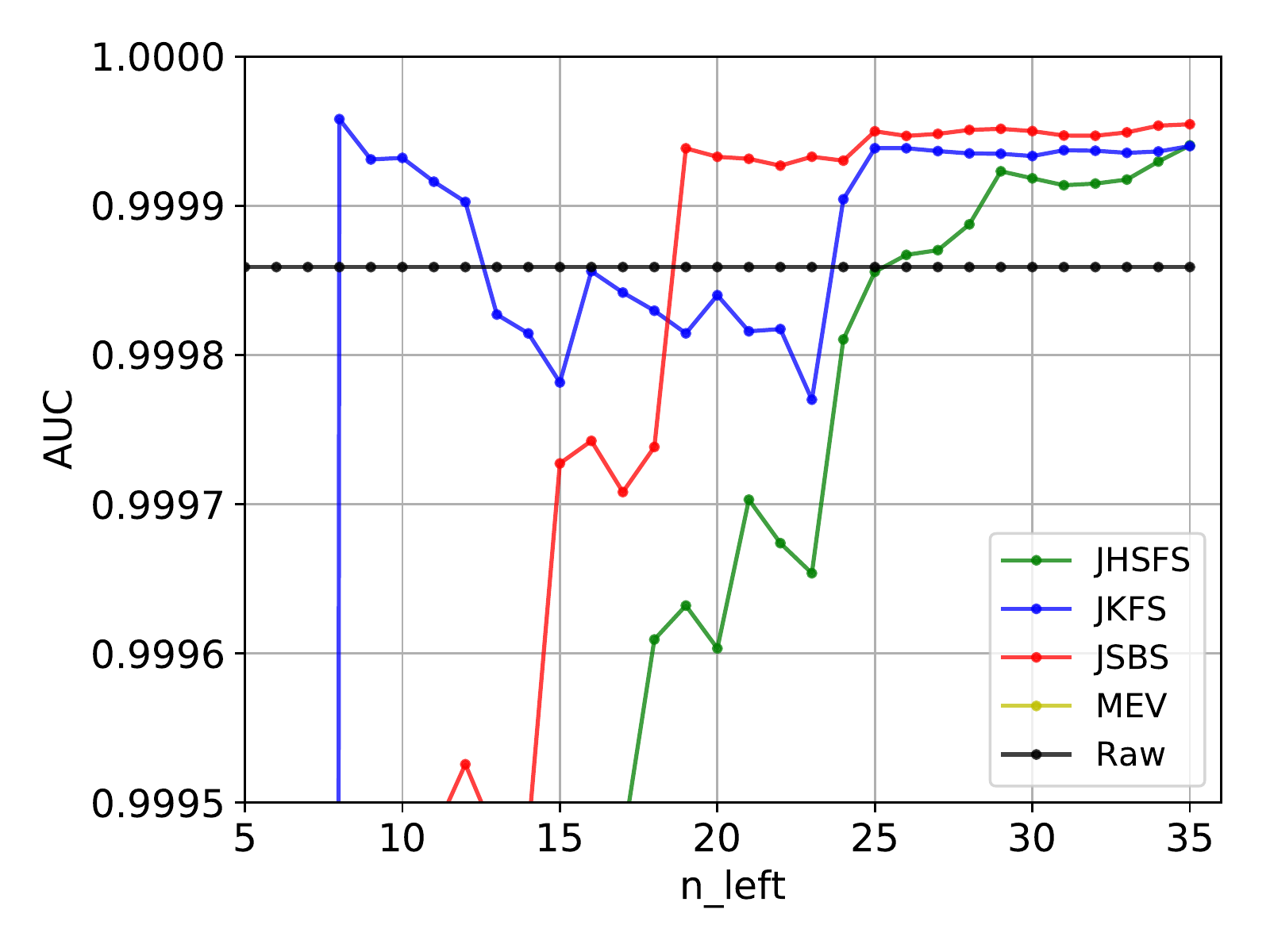}}
	\caption{Results of Buddigntonite deposit detection in the Curprite dataset
	compared to the scenario when no band selection is used (black line).
	Panel~(a) presents an Area Under Curve (AUC) as a function of the number of
	selected bands $n_{\text{left}}$ for three methods of band selection based
	on HOMC (JSBS, JKFS, JHSFS) and for band selection with
	the MEV algorithm. Beginning with values of $n_{\text{left}} = 8$ bands
	left, detection with HOMC-based methods outperforms the
	detection without band selection. The magnification of the vertical axis
	for this area is presented in the Panel~(b). Results illustrate the
	presence of `breaking points', discussed in the
	Section~\ref{sec::parran}.
	}\label{fig::AUC}
\end{figure}

\begin{figure}
	\subfigure[$n_{\text{left}} =
	10$]{\includegraphics[scale = 0.48]{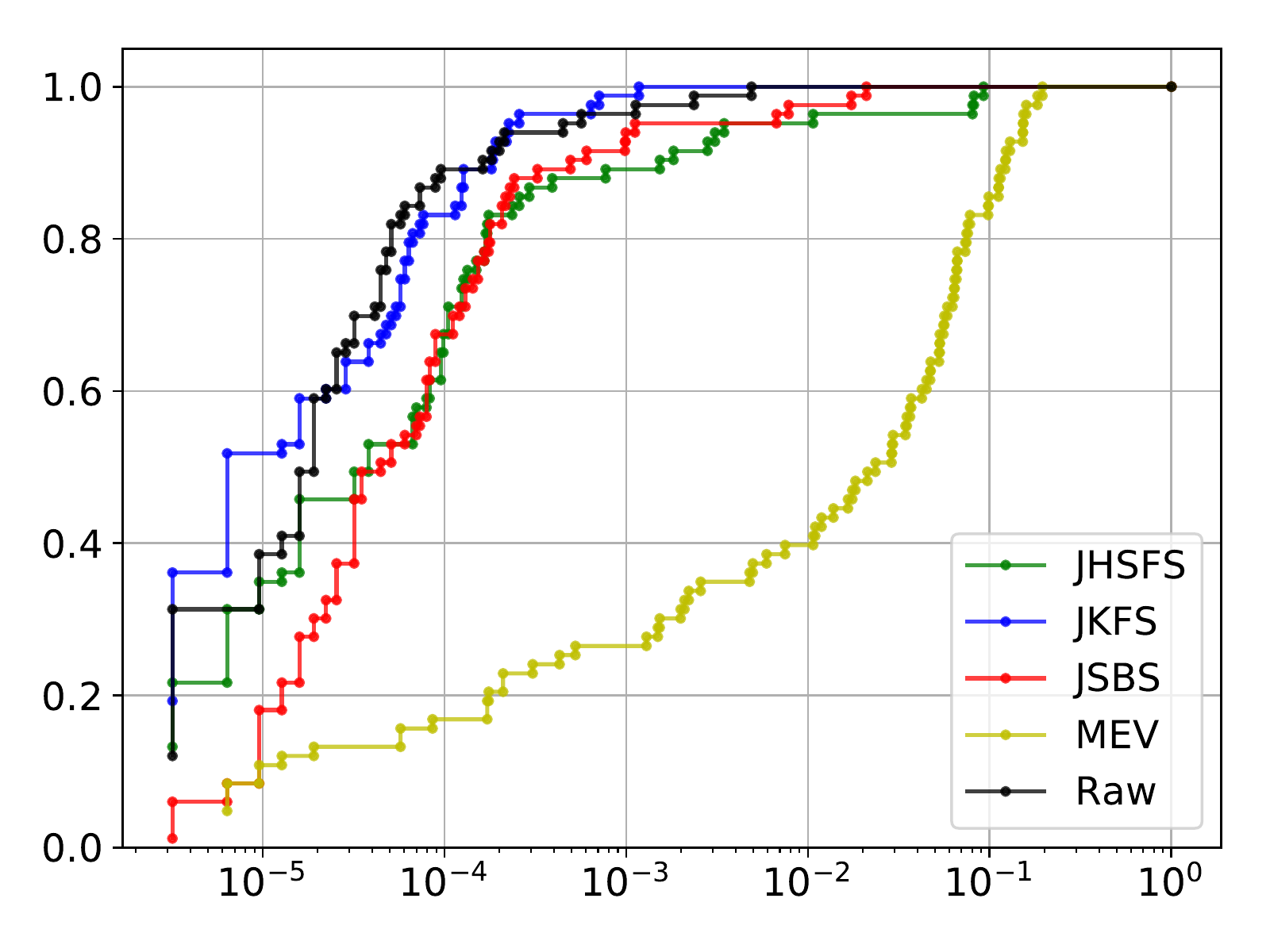}}
	\subfigure[$n_{\text{left}} =
	9$]{\includegraphics[scale = 0.48]{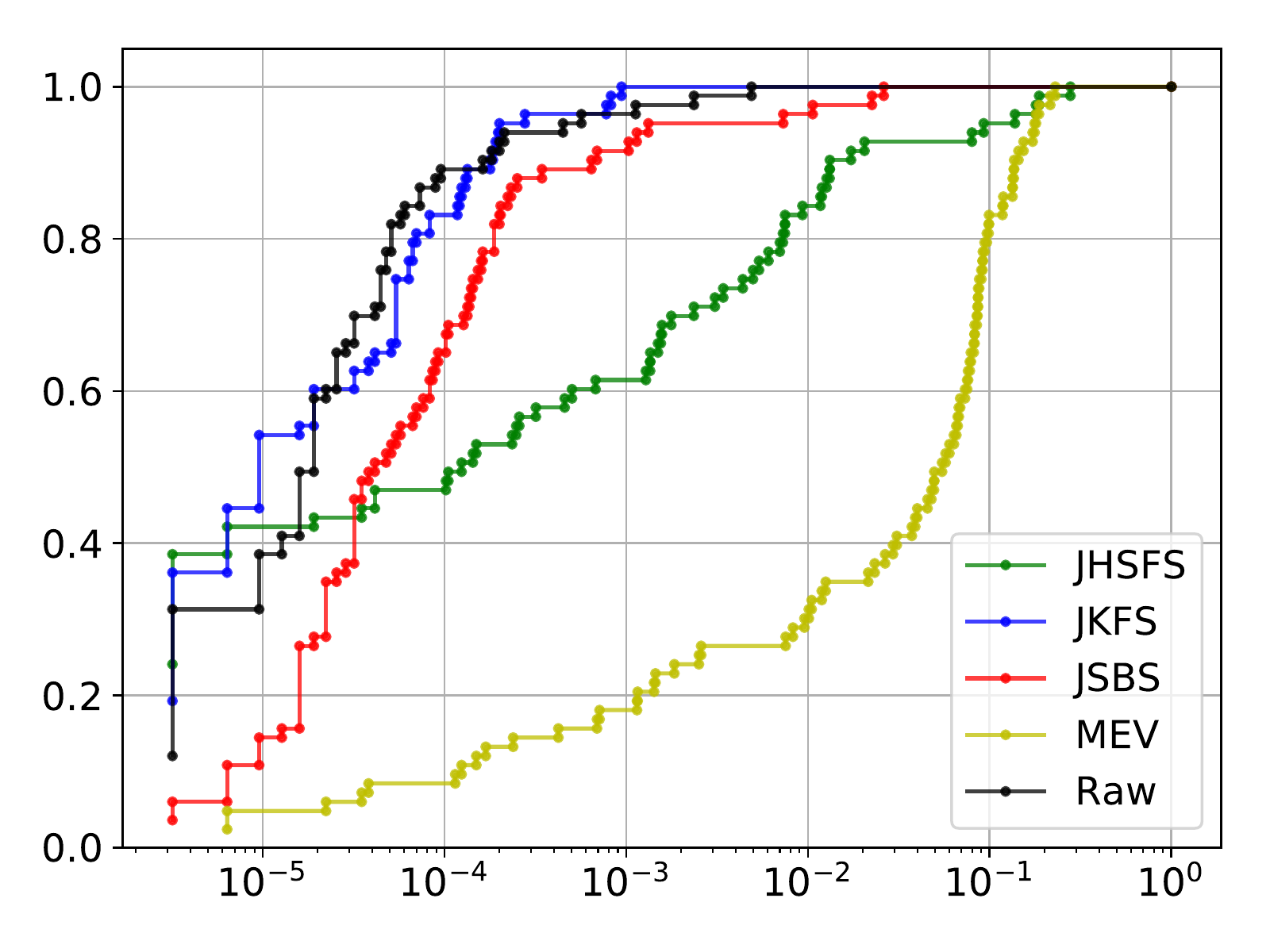}} \\
	\subfigure[$n_{\text{left}} =
	7$]{\includegraphics[scale = 0.48]{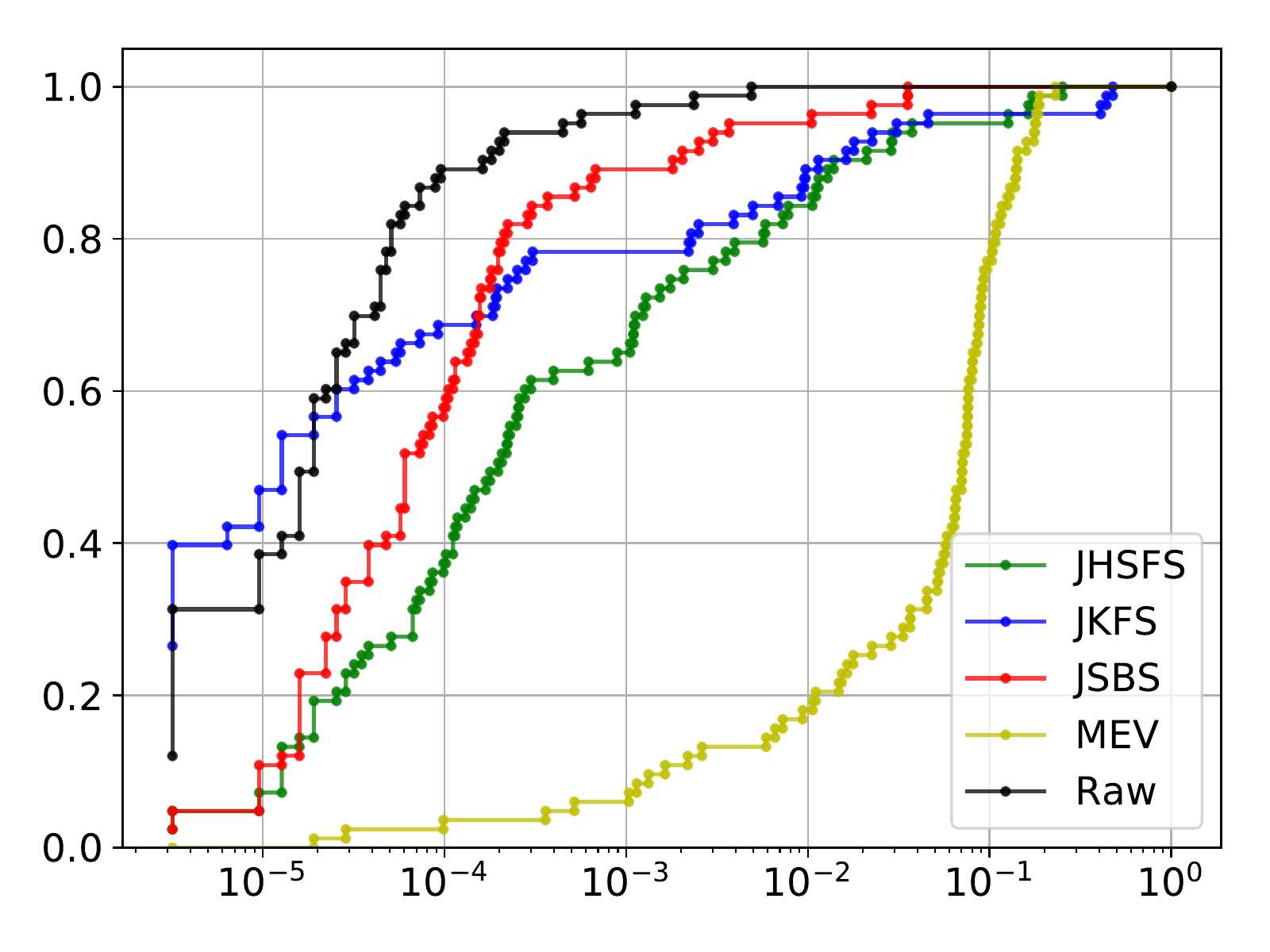}}
	\subfigure[$n_{\text{left}} =
	4$]{\includegraphics[scale = 0.48]{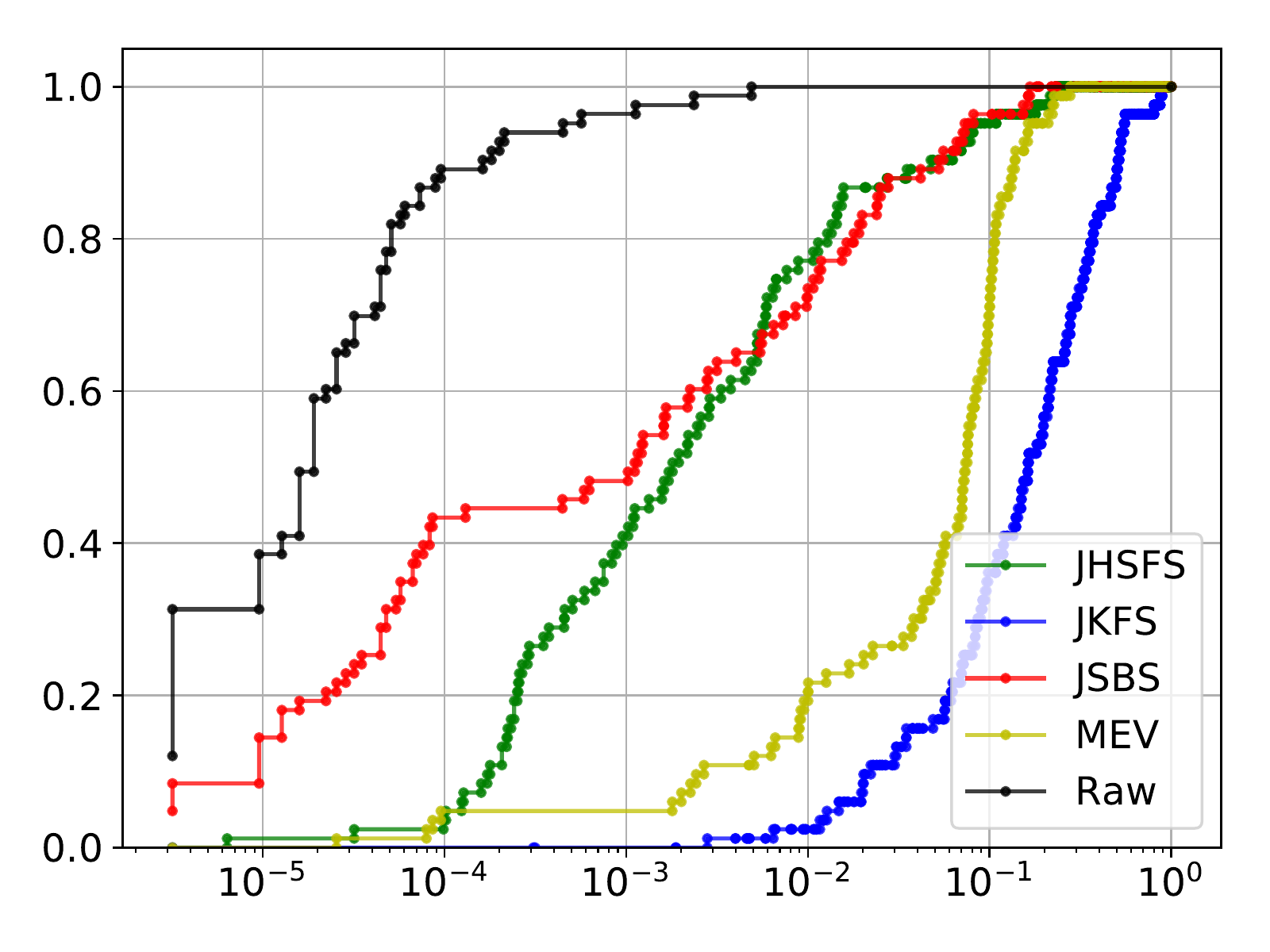}}
	\caption{An illustration of breakdowns in accuracy for band selection
	methods based on HOMC when the desired number of bands $n_{\text{left}}$
	becomes too low. Panel~(a) presents the scenario when the number of
	selected band $n_{\text{left}}$ is sufficient for all methods. Panel~(b)
	presents the breakdown in accuracy for JHSFS method (order $d=5$).
	Panel~(c) presents the breakdown in accuracy for JKFS method (order $d=4$).
	Panel~(d) presents the breakdown in accuracy for JSBS method (order
	$d=3$). Observe that each ROC curve loses its convexity at the breakdown 
	point.}\label{fig::c4c5break}
\end{figure}

\begin{figure}
	\subfigure[$n_{\text{left}} =
	14$]{\includegraphics[scale = 0.48]{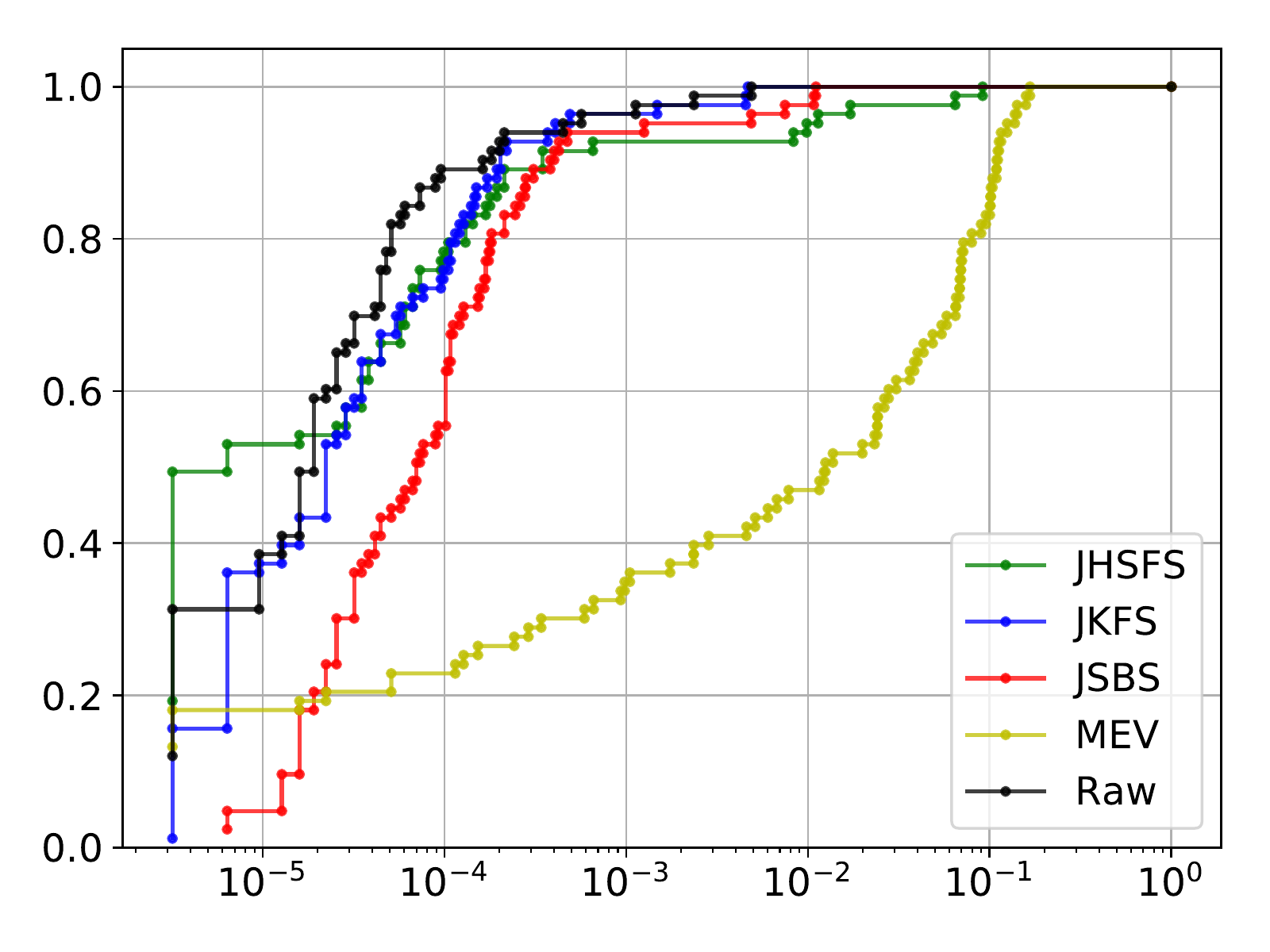}}
	\subfigure[$n_{\text{left}} =
	13$]{\includegraphics[scale = 0.48]{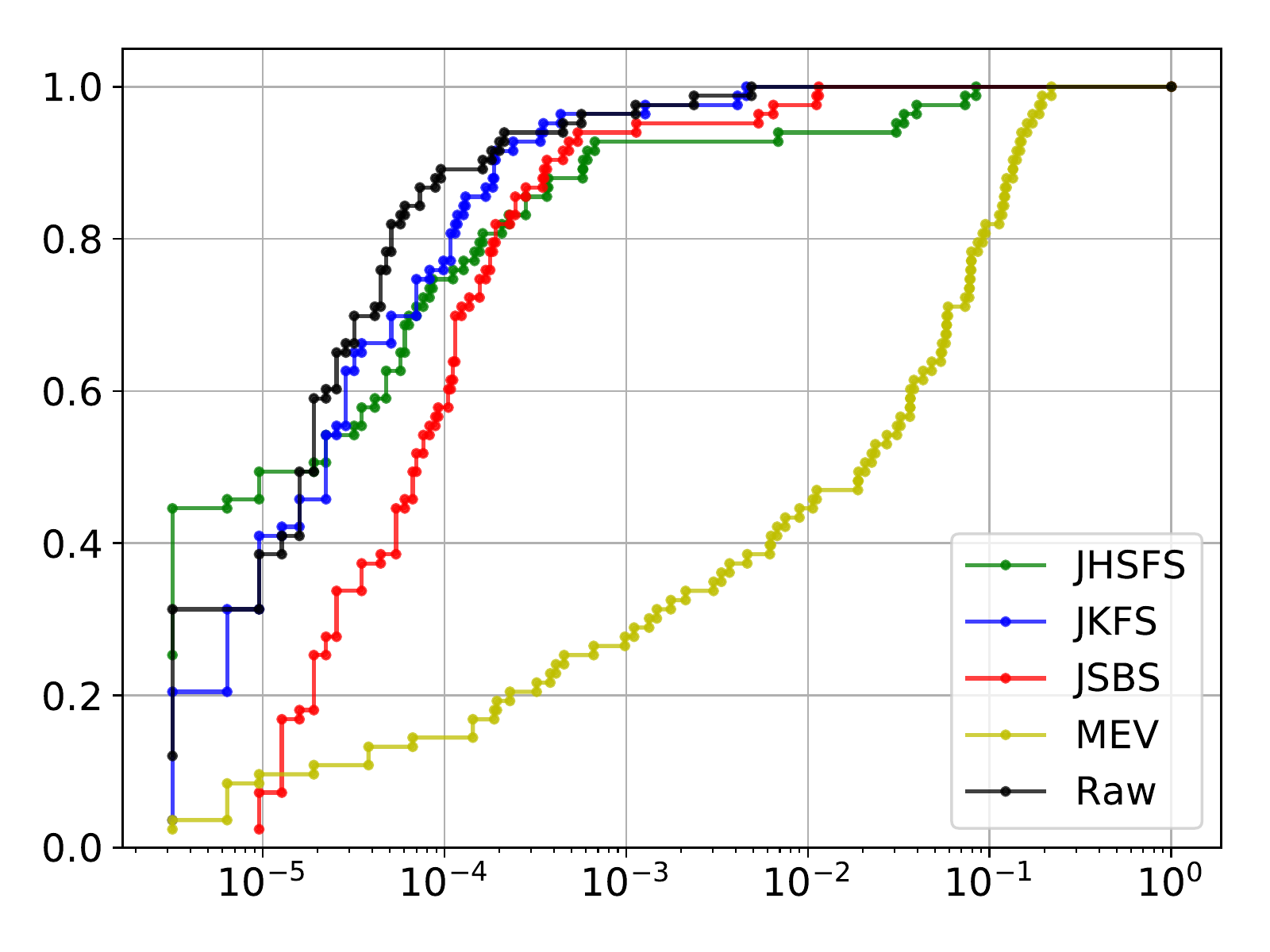}}
	\caption{An illustration of superior performance of the JSHFS if very low false positive
		probability is required.}\label{fig::c5worrks}
\end{figure}

\subsection{Results and discussion}
The performance of target detector using evaluated methods of band selection is presented
in the Figure~\ref{fig::AUC}. The detector using HOMC-based methods outperforms
the detector based on MEV and for the number of bands greater then
$n_{\text{left}} = 7$, it also usually outperforms the detector that does not
use band selection. Note that if a band selection method achieves the same
score as detector applied on all bands it can already be viewed as a success, as
we reduce the amount of working data. The improved result is thus an additional advantage. 

From an application point of view, the best performance is achieved by the $d = 4$ (JKFS) method proposed in this work with $n_{\text{left}} = 8$; it achieves highest score at with the lowest number of retained bands, thus maximizing the detection performance while minimizing the data volume.
As discussed in the Section~\ref{sec::parran}, for large values of the parameter
$n_{\text{left}}$ all methods behave in a similar fashion. However, for small values of the parameter
$n_{\text{left}}$ we can observe sharp breakdowns in the performance of each method: at $n_{\text{left}} = 4$ bands for the order $d =
3$ (JSBS), at $n_{\text{left}} = 7$ bands for the order $d = 4$ (JKFS) and
at $n_{\text{left}} = 9$ bands for the order $d = 5$ (JHSFS). A more detailed illustration of these breakdowns is provided in Fig.~\ref{fig::c4c5break}. Note that experimental results are consistent with the theoretical lower limit of the $n_{\text{left}}$ parameter, discussed in Section~\ref{sec::parran}.
As presented in the Fig.~\ref{fig::c5worrks}, when the desired level of false positive rate (FPR) is low, the JHSFS algorithm significantly outperforms other methods, which corresponds to the assumption that it is more sensitive to outliers.

Methods of band selection based on HOMC are a promising approach to
hyperspectral small target detection. Our results show that they allow to
select a small subset of relevant bands that maintains or improves the
performance of the detector while reducing the dimensionality of the data by up
to $84\%$. They also significantly outperform standard methods of band
selection such as MEV.

An interesting observation regards the performance drop of evaluated methods,
when the number of selected bands is too low (the existence of `breaking
points'). This means that HOMC-based methods have limits to achievable
dimensionality reduction.
These breakdowns in performance correspond to a phase transition-like behaviour,
as presented in the Figure~\ref{fig::c4c5break}, which corresponds to the  changes in shape of the ROC curve from convex to non-convex.

Results presented in the Figure~\ref{fig::AUC}, Panel~(b) indicate that while
HOMC-based methods can achieve superior performance for small number of bands,
some variability of performance can be observed. This
may be caused by estimation error of higher order statistics. However, we
observe that the performance of a reference approach using MEV is unstable in a
significantly larger range.

As presented in the Fig.~\ref{fig::c5worrks}, HOMC are
especially effective when the desired level of false positive rate (FPR) is
low. While this sensitivity seems to be correlated with the cumulant order, we
decided to evaluate only cumulants up to the  $5$\textsuperscript{th} order.
Since the theoretical lower limit of the number of selected bands for the
methods based on $6$\textsuperscript{th} order cumulant  is approximately
$n_{\text{left}}=16$, and  the estimation error is higher than for JSHFS. We
expect that improvement for HOMC can be significant for
larger hyperspectral images with very high number of pixels. This requires only
a large amount of collected data (which can be performed automatically, e.g.
with satellite passes) and does not require any manual annotation, which can be
an expensive process. What is also important, computation of HOMC for large
datasets can be effectively implemented \cite{domino2018efficient}, which gives
the potential to implement HOMC-based
band selection methods for very large images.

\section{Conclusions}

Band selection based on Higher Order Multivariate Cumulants (HOMC) seems to be an effective and promising approach for small target detection in hyperspectral
	images. We have observed that HOMC are more sensitive to
	tails in multivariate distribution. In our opinion, this property corresponds has a major impact on their superior performance. On the other hand, HOMC-based methods may have high estimation error, which may cause some instability when the number of desired bands is low.

In this work we proposed a uniform derivation of $d$-order
	cumulant-based methods. Methods such as JSBS and JKFS are special cases of
	the proposed derivation. We also proposed a new method of band selection,
	called Joint Hyper Skewness Feature Selection (JHSFS). We have shown that
	in scenarios when the desired False Positive Rate (FPR) is low, the JHSFS
	can outperform both the JSBS and JKFS.

During the experimental evaluation we have observed that the performance
	of hyperspectral target detection using HOMC-based band selection drops
	sharply when the desired number of bands is too low. We noticed that these
	`breaking points' are correlated with the percentage of off diagonal
	elements
	in a $d$-mode cumulant tensor. Hence each of HOMC-based
	methods has a minimum required number of selected bands. We provide a
	theoretical
	justification for this fact, by observing that off diagonal elements of
	$d$-order
	cumulant's tensor measure a mutual correlation of $d$\textsuperscript{th}
	feature.

\subsubsection*{Acknowledgments}
This work was partially supported by the National Science Centre, Poland—
project number 2014/15/B/ST6/05204. Authors would like to thank Adam
Glos for his assistance in implementation of the described method.

\bibliographystyle{plain}
\bibliography{cumulantbandsel}

\end{document}